\documentclass{article} 
\usepackage{iclr2026_conference,times}

\usepackage{amsmath,amsfonts,bm}

\def\eqref#1{equation~\ref{#1}}

\def\1{\bm{1}}

\def\ra{{\textnormal{a}}}

\def\rx{{\textnormal{x}}}

\def\rva{{\mathbf{a}}}

\def\erva{{\textnormal{a}}}

\def\ervx{{\textnormal{x}}}

\def\rmA{{\mathbf{A}}}

\def\vmu{{\bm{\mu}}}
\def\vtheta{{\bm{\theta}}}
\def\va{{\bm{a}}}

\def\ve{{\bm{e}}}

\def\vx{{\bm{x}}}

\def\eva{{a}}

\def\mA{{\bm{A}}}

\def\mH{{\bm{H}}}
\def\mI{{\bm{I}}}
\def\mJ{{\bm{J}}}

\def\mX{{\bm{X}}}

\def\mSigma{{\bm{\Sigma}}}

\DeclareMathAlphabet{\mathsfit}{\encodingdefault}{\sfdefault}{m}{sl}
\SetMathAlphabet{\mathsfit}{bold}{\encodingdefault}{\sfdefault}{bx}{n}
\newcommand{\tens}[1]{\bm{\mathsfit{#1}}}
\def\tA{{\tens{A}}}

\def\tX{{\tens{X}}}

\def\gG{{\mathcal{G}}}

\def\sA{{\mathbb{A}}}
\def\sB{{\mathbb{B}}}

\def\sS{{\mathbb{S}}}

\def\emA{{A}}

\newcommand{\etens}[1]{\mathsfit{#1}}

\def\etA{{\etens{A}}}

\newcommand{\E}{\mathbb{E}}

\newcommand{\R}{\mathbb{R}}

\newcommand{\KL}{D_{\mathrm{KL}}}
\newcommand{\Var}{\mathrm{Var}}

\newcommand{\Cov}{\mathrm{Cov}}

\newcommand{\normltwo}{L^2}
\newcommand{\normlp}{L^p}

\newcommand{\parents}{Pa} 

\usepackage{hyperref}
\usepackage{url}

\usepackage{graphicx}%
\usepackage{multirow}%
\usepackage{amsmath,amssymb,amsfonts}%
\usepackage{amsthm}%
\usepackage{mathrsfs}%
\usepackage[title]{appendix}%
\usepackage{xcolor}%
\usepackage{textcomp}%
\usepackage{manyfoot}%
\usepackage{booktabs}%
\usepackage{algorithm}%
\usepackage{algorithmicx}%
\usepackage{algpseudocode}%
\usepackage{listings}%

\usepackage{natbib}
\usepackage{subcaption}

\usepackage{colortbl}

\definecolor{cb_orange}{RGB}{213,94,0}
\definecolor{cb_green}{RGB}{34,136,51}
\definecolor{cbgreen}{RGB}{34,136,51}
\definecolor{sky_blue}{RGB}{204, 238, 255}
\definecolor{cb_purple}{RGB}{170, 51, 119}
\definecolor{cb_red}{RGB}{204, 51, 17}
\definecolor{cb_blue}{RGB}{0, 119, 187}
\definecolor{mydarkblue}{rgb}{0,0.08,0.45}
\definecolor{forestgreen}{RGB}{34,139,34}
\definecolor{periwinkle}{rgb}{0.8, 0.8, 1.0}
\definecolor{royalazure}{rgb}{0.0, 0.22, 0.66}
\definecolor{royalblue}{rgb}{0.0, 0.14, 0.4}
\definecolor{richlilac}{rgb}{0.71, 0.4, 0.82}

\usepackage{bbding}
\usepackage{tcolorbox}
\usepackage{makecell}

\title{SSM Meets Video Diffusion Models:\\Efficient Long-Term Video Generation with Structured State Spaces}

\author{Yuta Oshima, Shohei Taniguchi, Masahiro Suzuki \& Yutaka Matsuo 
\\
The University of Tokyo\\
7-3-1, Hongo, Bunkyo-ku, Tokyo, Japan. \\
\texttt{\{yuta.oshima, taniguchi, masa, matsuo\}@weblab.t.u-tokyo.ac.jp} 
}

\begin{document}

\maketitle

\begin{abstract}
Given the remarkable achievements in image generation using diffusion models, the research community has shown increasing interest in extending these models to video generation. 
Recent diffusion models for video generation have predominantly utilized attention layers to extract temporal features. 
However, attention layers are limited by their computational cost, which increases quadratically with sequence length. 
This limitation poses significant challenges when generating longer video sequences using diffusion models. 
To overcome these challenges, we propose to leverage state-space models (SSMs) as temporal feature extractors. SSMs (e.g., Mamba) have recently garnered attention as promising alternatives owing to their linear-time memory and time consumption relative to the sequence length. 
Employing SSMs to capture temporal dependencies in video generation enables significantly higher generative performance at the same computational cost (e.g., memory usage, inference time) compared to attention-based methods, particularly for long-term sequences. 
For various model sizes, we comprehensively evaluated multiple long-term video datasets: MineRL Navigate, GQN-Mazes, and CARLA-Town01.
For 256-frame video sequences, SSM-based models incur lower computational cost to achieve the same Fréchet Video Distance as attention-based models. 
Furthermore, the ablation study shows that when using SSMs for temporal modeling, incorporating bidirectionality and selective scans enhances video generation performance. 
Our code is available at \url{https://github.com/shim0114/SSM-Meets-Video-Diffusion-Models}.
\end{abstract}

\section{Introduction}
Research on video generation using diffusion models~\citep{origin-diffusion, origin-diffusion2, ddpm} is at the cutting edge of deep generative models. 
The success of image generation using diffusion models, notably Denoising Diffusion Probabilistic Models (DDPMs)~\citep{ddpm}, has sparked a surge in studies on applying diffusion models to video generation. 
This trend is exemplified by the emergence of video diffusion models (VDMs)~\citep{vdm}. 
Leveraging the substantial representational capacity inherent in diffusion models, these models have shown impressive performance in modeling the intricate nature of video content~\citep{imagenvideo, makeavideo, sora}.

\begin{figure}[t]
    \centering
    \includegraphics[width=\linewidth]{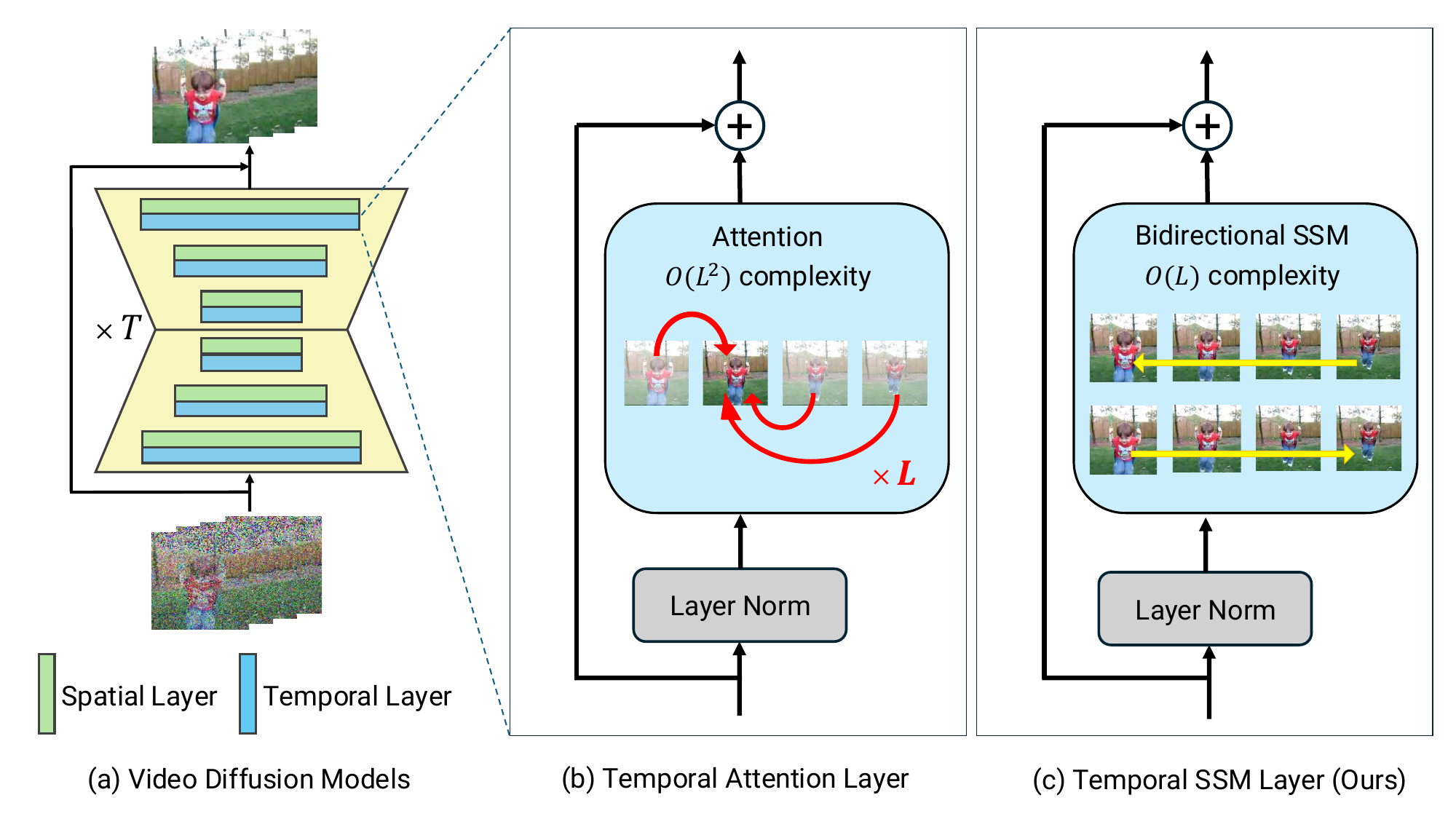}
    \caption{(a) U-Net based video diffusion models comprising spatial layers and temporal layers. (b) Conventional approaches using attention mechanisms in temporal layers, leading to quadratic growth in memory allocation and computational time as sequence length $L$ increases. (c) We propose using SSMs in temporal layers, where memory allocation and computational time increase only linearly with sequence length $L$.}
    \label{fig1}
\end{figure}

However, research on diffusion-model-based video generation faces significant challenges of computational complexity with respect to the video sequence length. 
In diffusion models for video generation, attention mechanisms~\citep{attention} have been employed to capture temporal relationships~\citep{vdm, makeavideo, imagenvideo, magicvideo}. 
In previous studies on diffusion models for videos, such as VDMs, temporal attention layers were added after spatial attention layers within the architecture of diffusion models for image generation to capture temporal relationships across video frames, as shown in \autoref{fig1}(a)(b). 
However, the computational demands of attention layers, which scale quadratically with sequence length, pose substantial challenges for extending these models to handle longer sequences.
While some recent models, such as Sora~\citep{sora}, have begun to tackle longer video generation, practical memory and computational time constraints often limit the sequence lengths that can be produced. 
Consequently, many state-of-the-art methods~\citep{wang2023lavie, chen2024videocrafter2} still focus on short-term video generation, often producing only 16 frames per inference, corresponding to just two seconds of footage at 8 fps.

To overcome the challenge of computational complexity associated with increasing video sequence length, we propose leveraging SSMs as temporal feature extractors (\autoref{fig1}(c)).
Recently, state-space models (SSMs)~\citep{s4, s4d, s5}, particularly Mamba~\citep{mamba}, have been identified as promising alternatives to attention mechanisms. 
In contrast to attention mechanisms, SSMs can handle sequential data with linear complexities; therefore, they are expected to overcome the fundamental limitations of attention-based models in several sequence modeling tasks.
In other words, by leveraging SSMs to capture temporal relationships in video generation models, one can potentially achieve higher generative quality within the same computational budget (e.g., memory usage, inference time) than attention-based methods, especially in long-term video generation settings, owing to their computational efficiency.


A key finding of this study is that architectural choices within the temporal SSM layer are crucial: bidirectionality~\citep{bi-lstm, visionmamba} is needed to capture temporal dynamics that depend on both past and future frames, and selective scans~\citep{mamba} enable input-adaptive state updates across frames. Naively replacing temporal attention with a unidirectional or non-selective SSM substantially underperforms the original attention-based VDM, whereas combining both choices yields the gains reported below. We defer the detailed mechanism to Section~\ref{ablation_study}.

First, we conducted experiments on the widely used long-term video dataset MineRL Navigate~\citep{minerl,cwvae} using sequences of 16, 64, and 256 frames, and on GQN-Mazes~\citep{eslami2018neural, cwvae} and CARLA Town01~\citep{dosovitskiy2017carla, fdm} using 256 frames. 
We compared SSM-based models and attention-based models of various sizes in terms of computational cost and video generation quality. 
Under the 256-frame setting, we observed that SSM-based models achieve a lower Fréchet Video Distance (FVD), a metric of video generation quality, than their attention-based counterparts under the same memory and time constraints, and remain competitive for 16- and 64-frame sequences. 
These findings suggest that, particularly for long-term video generation, SSM-based models can reduce the computational cost to achieve generation quality comparable to attention-based approaches. 
Furthermore, our ablation studies revealed that adopting bidirectionality and selective scans in the SSM layers can further improve FVD.

\section{Background}

\subsection{Denoising Diffusion Probabilistic Models (DDPMs)}
In diffusion models, the forward process progressively diminishes the original data signal, $\mathbf{x}_0$, by gradually introducing Gaussian noise as the diffusion time $t$ advances.
This sequence of transformations leads $\mathbf{x}_0$ to converge to pure Gaussian noise, represented as $\mathbf{x}_T \sim \mathcal{N}(\mathbf{x}_T; \mathbf{0}, \mathbf{I})$, at time $T$. 
In this study, $t$ is treated as a discrete integer within the range $[0, T]$, although some studies have considered $t$ to be a continuous variable~\cite{scorebased, vardm}.The forward process is governed by the following Markov process:
\begin{gather}
q(\mathbf{x}_{1:T} | \mathbf{x}_0) = \prod_{t=1}^{T} q(\mathbf{x}_t | \mathbf{x}_{t-1}), \\
q(\mathbf{x}_t | \mathbf{x}_{t-1}) = \mathcal{N} (\mathbf{x}_t ; \sqrt{\alpha_t}\,\mathbf{x}_{t-1}, (1 - \alpha_t)\mathbf{I}),
\end{gather}
where $\alpha_t = 1 - \sigma_t^2 \in (0,1)$ controls the per-step signal retention, and the noise schedule $\{\sigma_t\}_{t=1}^{T}$ satisfies $0 < \sigma_1 < \cdots < \sigma_{T-1} < \sigma_T < 1$.
Equivalently, the marginal admits the closed form $q(\mathbf{x}_t|\mathbf{x}_0) = \mathcal{N}(\mathbf{x}_t; \sqrt{\bar{\alpha}_t}\,\mathbf{x}_0, (1-\bar{\alpha}_t)\mathbf{I})$ with $\bar{\alpha}_t = \prod_{i=1}^{t}\alpha_i = \prod_{i=1}^{t}(1-\sigma_i^2)$.

The generation process in the diffusion models is the reverse process. This process starts with pure Gaussian noise $\mathbf{x}_T \sim \mathcal{N}(\mathbf{x}_T; \mathbf{0}, \mathbf{I})$ and gradually reconstructs the data towards the original \( \mathbf{x}_0 \). During the reverse process, each step \( p_{\theta}(\mathbf{x}_{t-1}|\mathbf{x}_{t}) \) is modeled using a neural network parameterized by \( \theta \). 
\begin{gather}
p_{\theta}(\mathbf{x}_{1:T})=p(\mathbf{x}_T)\prod_{t=1}^{T} p_{\theta}(\mathbf{x}_{t-1} | \mathbf{x}_t), \\
p_{\theta}(\mathbf{x}_{t-1} | \mathbf{x}_t) = \mathcal{N} (\mathbf{x}_{t-1} ; \boldsymbol{\mu}_{\theta}(\mathbf{x}_t,t), \boldsymbol{\Sigma}_{\theta}(\mathbf{x}_t,t)), \\
p(\mathbf{x}_T) = \mathcal{N}(\mathbf{x}_T;\mathbf{0},\mathbf{I}).
\end{gather}
Typically, $\boldsymbol{\Sigma}_{\theta}$ is set as an untrainable, time-dependent constant, \( \boldsymbol{\Sigma}_{\theta}(\mathbf{x}_t, t) = \sigma_t \mathbf{I} \). 
Additionally, with a change in the parameterization of \( \boldsymbol{\mu}_{\theta} \), the reverse process \( p_{\theta}(\mathbf{x}_{t-1} | \mathbf{x}_t) \) can be expressed as follows:
\begin{equation}
p_{\theta}(\mathbf{x}_{t-1} | \mathbf{x}_t) = \mathcal{N} \left( \mathbf{x}_{t-1} ; \frac{1}{\sqrt{\alpha_{t}}}\left(\mathbf{x}_{t} - \frac{\sigma_{t}^2}{\sqrt{1-\bar\alpha_{t}}}\boldsymbol{\epsilon}_{\theta}(\mathbf{x}_t, t)\right), \sigma_t \mathbf{I} \right),
\label{eq:reverse-process}
\end{equation}
where \(\bar{\alpha}_t\) is as defined above.
Term \( \boldsymbol{\epsilon}_{\theta}(\mathbf{x}_t, t) \) represents a function that predicts the noise from noisy data \( \mathbf{x}_t \). This parameterization results in the following objective function for the DDPM:
\begin{equation}
\mathbb{E}_{\mathbf{x}_0, \boldsymbol{\epsilon},t } \left[ \left\| \boldsymbol{\epsilon} - \boldsymbol{\epsilon}_{\theta}(\mathbf{x_t},t) \right\|^2_2 \right],
\end{equation}
where \( \mathbf{x}_t = \sqrt{\bar\alpha_t}\mathbf{x}_0+\sqrt{1-\bar\alpha_t}\boldsymbol{\epsilon} \).
Whereas numerous parameterizations are recognized, such as predicting the observed data \( \mathbf{x}_0 \) from its noisy counterpart \( \mathbf{x}_t \) or \( \mathbf{v} \)-prediction~\citep{vdm, distillation, vardm}, we utilized the \( \boldsymbol{\epsilon} \)-prediction.

In terms of diffusion model architecture, 2D U-Net~\citep{unet} is commonly used for image data. 
In 2D U-Net-based models, spatial attention layers are incorporated between the convolutional layers. 
These spatial attention layers enhance the ability to focus on relevant spatial features, thereby improving the quality of the generated images.
Although we adopted a U-Net-based architecture for the diffusion model, Diffusion Transformers (DiT)~\citep{dit} explored architectures based on Vision Transformers (ViT)~\citep{vit}. 

\subsection{Architectures for Video Diffusion Models}
For video generation, diffusion models must encapsulate both spatial and temporal features across frames.
Although DDPMs typically comprise a U-Net and spatial attention layers, their capabilities are predominantly limited to spatial feature capture.
To address this limitation, Video Diffusion Models (VDMs)~\citep{vdm} were introduced as an initial attempt at video generation. 
By incorporating mechanisms to capture temporal dynamics within DDPMs, VDMs enhance their capability of capturing temporal features (\autoref{fig1} (a)(b)).
Temporal attention layers are commonly used in video generation diffusion models, such as VDMs, to leverage time-series dependencies. However, temporal attention requires memory and time proportional to the square of the sequence length, which limits the maximum length of video sequences that can be generated in a single sampling loop. 
In this study, we adopted VDMs as a baseline to explore the existing challenges and potential improvements in video generation using diffusion models.

\section{Method}

\begin{figure}[t]
    \centering
    \includegraphics[width=\linewidth]{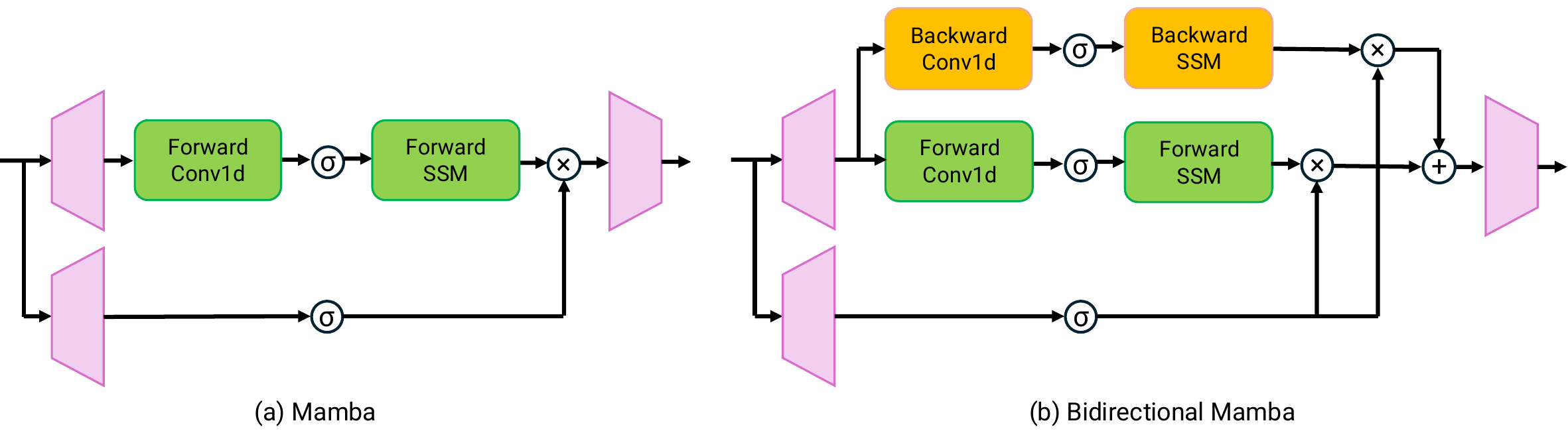}
    \caption{Architectural comparison of Mamba and bidirectional Mamba. Layer normalization~\citep{layernorm} and a skip connection~\citep{resnet} are omitted for simplicity.}
    \label{mamba_bidirectional}
\end{figure}

In this section, we propose the architecture of a temporal SSM layer for use in video diffusion models. 
Recently, SSMs have emerged as promising alternatives to attention mechanisms, offering linear memory and time complexity with respect to sequence length~\citep{hippo, s4, s4d, s5, mamba}. 
First, we review recent advancements in SSMs from previous studies, followed by a detailed description of the proposed temporal SSM layer architecture for video-generation diffusion models.

\subsection{State-space Models}
Unlike temporal attention commonly used in video diffusion models, state-space models (SSMs) enable processing of time series with spatial complexity proportional to the sequence length. 
Recent studies have proposed SSMs that can process inputs in parallel, unlike recurrent models such as recurrent neural networks (RNNs)~\citep{rnn}.
SSMs are widely used as sequence models that define a mapping from one-dimensional input signals \( u(t) \in \mathbb{R} \) to one-dimensional output signals \( y(t) \in \mathbb{R}\), with \( \mathbf{s}(t) \in \mathbb{R} ^ N\) representing the hidden state.
The continuous-time processes are formulated as follows:

\begin{equation}
\begin{aligned}
    \dot{\mathbf{s}}(t) &= \mathbf{A}\mathbf{s}(t) + \mathbf{B}u(t), \\
    y(t) &= \mathbf{C}\mathbf{s}(t) + Du(t), \\
\end{aligned}
\label{ssm-continuous}
\end{equation}
where \(\mathbf{A}\in \mathbb{R}^{N\times N}\), \(\mathbf{B}\in \mathbb{R}^{N\times 1}\), and \(\mathbf{C}\in \mathbb{R}^{1\times N}\) are the diagonal state, input, and output matrices, respectively, and \(D\in \mathbb{R}\) is the direct (skip) path from the input to the output.
In practice, following the Mamba implementation~\citep{mamba}, $D$ is realized as a learnable per-channel skip connection applied independently in both the forward and backward branches of the bidirectional layer; we omit $D$ from Algorithm~\ref{alg:temporal_ssm} for brevity.
To apply SSMs to real-world data, the SSMs formulated in \autoref{ssm-continuous} are converted into discrete versions utilizing the zero-order hold (ZOH) method~\citep{s4d} as follows:

\begin{equation} 
\begin{aligned}
\mathbf{s}_{k} &= \mathbf{\bar{A}}\mathbf{s}_{k-1} + \mathbf{\bar{B}}u_{k}, \\
y_{k} &= \mathbf{C}\mathbf{s}_{k} + Du_k,
\end{aligned}
\label{ssm-discrete}
\end{equation}
where \(\mathbf{\bar{A}} = \exp(\mathbf{\Delta}\mathbf{A})\), \(\mathbf{\bar{B}} = (\mathbf{\Delta}\mathbf{A})^{-1}\bigl(\exp(\mathbf{\Delta}\mathbf{A})-\mathbf{I}\bigr)\cdot\mathbf{\Delta}\mathbf{B}\), with \(\mathbf{\Delta}\) being the timescale parameter for ZOH.
In contrast to previous SSMs such as S4D~\citep{s4d}, a diagonal-state-matrix variant of S4 limited to linear time-invariant (LTI) systems, Mamba~\citep{mamba} introduces a selective scan mechanism (S6).
In S6, the parameters \(\mathbf{B} \in \mathbb{R} ^ {B \times L \times N}\), \(\mathbf{C} \in \mathbb{R} ^ {B \times L \times N}\), and \(\mathbf{\Delta} \in \mathbb{R} ^ {B \times L \times D}\) are dependent on input \(\mathbf{u} \in \mathbb{R} ^ {B \times L \times D}\).
This enables the model to selectively discard irrelevant information while retaining essential information over extended periods.
Using an efficient scan algorithm, S6 can perform parallel computation with linear complexity relative to the sequence length.

\begin{algorithm}[t]
\caption{Temporal SSM Layer}
\label{alg:temporal_ssm}
\renewcommand{\arraystretch}{1.2} 
\begin{algorithmic}[1]
\Require input tensor \(\mathbf{T} \hfill \textcolor{teal}{\in \mathbb{R}^{(B \times H \times W) \times L \times C}}\)  

\Ensure output tensor \(\mathbf{U} \hfill \textcolor{teal}{\in \mathbb{R}^{(B \times H \times W) \times L \times C}}\)  

\State \(\mathbf{X} \leftarrow \text{Linear}^{\mathbf{X}}(\mathbf{T}) \hfill \textcolor{teal}{\in \mathbb{R}^{(B \times H \times W) \times L \times 2C}}\)  

\State \(\mathbf{Z} \leftarrow \text{Linear}^{\mathbf{Z}}(\mathbf{T}) \hfill \textcolor{teal}{\in \mathbb{R}^{(B \times H \times W) \times L \times 2C}}\)

\State \textbf{for} $o$ in \{$forward$, $backward$\} \textbf{do} 

\State ~~~~\(\mathbf{X'}_{o} \leftarrow \text{SiLU}(\text{Conv1D}_{o}(\mathbf{X}))   \hfill \textcolor{teal}{\in \mathbb{R}^{(B \times H \times W) \times L \times 2C}}\)  

    \State ~~~~\(\mathbf{B}_{o} \leftarrow \text{Linear}^{\mathbf{B}}_{o}(\mathbf{X'}_{o}) \hfill \textcolor{teal}{\in \mathbb{R}^{(B \times H \times W) \times L \times N}}\)  

\State ~~~~\(\mathbf{C}_{o} \leftarrow \text{Linear}^{\mathbf{C}}_{o}(\mathbf{X'}_{o})   \hfill \textcolor{teal}{\in \mathbb{R}^{(B \times H \times W) \times L \times N}}\)  

\State ~~~~\(\mathbf{\Delta}_{o} \leftarrow \text{Softplus}(\text{Linear}^{\mathbf{\Delta}}_{o}(\mathbf{X'}_{o})+\text{Parameter}^{\mathbf{\Delta}}_{o})\) 
\hfill \textcolor{teal}{\(\in \mathbb{R}^{(B \times H \times W) \times L \times 2C}\)}  

\State ~~~~\(\mathbf{\bar{A}}_{o}, \mathbf{\bar{B}}_{o} \leftarrow \text{discretize}(\mathbf{\Delta}, \mathbf{A}_{o}, \mathbf{B}_{o})  \hfill \textcolor{teal}{\in \mathbb{R}^{(B \times H\times W) \times L \times 2C}}\)

\State ~~~~\(\mathbf{X''}_{o} \leftarrow\text{SSM}_{o}(\mathbf{\bar{A}}_{o}, \mathbf{\bar{B}}_{o}, \mathbf{C}_{o})(\mathbf{X'}_{o})\)  
\hfill \textcolor{teal}{ \(\in \mathbb{R}^{(B \times H \times W) \times L \times 2C \times N}\) }

\State ~~~~\(\mathbf{Y}_{o} \leftarrow \mathbf{X''}_{o} \odot \text{SiLU}(\mathbf{Z})   \hfill \textcolor{teal}{\in \mathbb{R}^{(B \times H \times W) \times L \times 2C}}\)  

\State \(\mathbf{U} \leftarrow \text{Linear}^{\mathbf{U}}(\mathbf{Y}_{forward} + \mathbf{Y}_{backward})\)
\hfill \textcolor{teal}{\(\in \mathbb{R}^{(B \times H \times W) \times L \times C}\)}

\State \Return \(\mathbf{U}\)
\end{algorithmic}
\end{algorithm}

\subsection{Bidirectional Mamba}
\label{sec:bidirectional-mamba}
A single SSM is inherently causal and captures only unidirectional transitions. To overcome this limitation, prior work on sequence and vision modeling has introduced bidirectional variants~\citep{bi-lstm, pretrainssm}: in particular, Vision Mamba~\citep{visionmamba} processes image-patch sequences with two parallel Mamba branches running in forward and backward directions, and similar multi-directional designs~\citep{diffussm, vmamba} have been shown to improve generation quality in the spatial domain.
\autoref{mamba_bidirectional} illustrates this bidirectional Mamba block; the forward and backward branches share the same input but use independent Conv1D and SSM parameters, and their outputs are summed before a final linear projection.

\subsection{Proposed Temporal SSM Layer}
\label{sec:temporal-ssm}
We propose incorporating SSMs into the \emph{temporal} layers of a video diffusion U-Net (\autoref{fig1}(c)), replacing the temporal self-attention while leaving the spatial layers unchanged. To our knowledge, this is the first study to apply a bidirectional Mamba block specifically to the temporal axis of a video diffusion model for long-term ($\geq 256$-frame) generation; while bidirectional Mamba has been used for spatial patches in image models~\citep{visionmamba, vmamba}, the temporal axis poses distinct challenges due to much longer sequence lengths and the strong cross-frame dependencies inherent to video data.

Concretely, we adopt the bidirectional Mamba block of~\citep{visionmamba} (\autoref{mamba_bidirectional}) and, following standard Mamba setups~\citep{mamba, visionmamba, vmamba}, set the linear expansion factor to $E=2$ and the internal state dimension to $N=16$, with SiLU~\citep{silu} as the activation function. The temporal SSM layers are alternated with the spatial layers in the U-Net (\autoref{fig1}(a)). Within each temporal layer, we reshape the input from $(B \times L) \times C \times H \times W$ to $(B \times H \times W) \times L \times C$ so that the SSM operates along the temporal axis, apply layer normalization~\citep{layernorm}, and then feed the result into the bidirectional Mamba block. Here $L$, $C$, $H$, and $W$ denote the sequence length, channel size, height, and width of the input, respectively. The full processing is summarized in Algorithm~\ref{alg:temporal_ssm}.
The state matrix $\mathbf{A}_{o} \in \mathbb{R}^{2C \times N}$ for each direction $o \in \{\text{forward}, \text{backward}\}$ is a learnable parameter that is \emph{not} input-dependent, following standard Mamba practice~\citep{mamba}; it is initialized using the S4D~\citep{s4d} initialization. In contrast, $\mathbf{B}_{o}$, $\mathbf{C}_{o}$, and $\mathbf{\Delta}_{o}$ are produced from the input by linear projections (lines~5--7 of Algorithm~\ref{alg:temporal_ssm}), realizing the selective scan mechanism (S6)~\citep{mamba}.

\section{Related Works}

\subsection{Diffusion Models for Video Generation}
The advent of diffusion models in image generation~\citep{ddpm, ldm} marked a turning point, with subsequent expansions to video distributions showing promising results~\citep{vdm, wang2023lavie, chen2024videocrafter2, sora, gdm2024veo}. 
Nevertheless, attention-based temporal layers still lead to quadratic growth in memory and computation with respect to sequence length, posing substantial challenges. 
To address this challenge, two main strategies have been proposed: one employs multiple specialized models to handle different video segments or subtasks—often through interpolation or super-resolution~\citep{makeavideo, imagenvideo, nuwaxl}—while the other extends the video autoregressively, using a single model conditioned on previously generated frames~\citep{mcvd, ramvid, fdm, qiu2023freenoise, wang2023gen, kim2024fifo}.
Both strategies reduce memory usage by restricting the temporal span each model processes, but their iterative nature inherently increases total computation time. 
In contrast, this study emphasizes architectural improvements rather than sampling schemes, aiming to model the entire video sequence without sacrificing time efficiency.

We also note that recent high-profile video diffusion systems such as LaVie~\citep{wang2023lavie}, VideoCrafter2~\citep{chen2024videocrafter2}, and NUWA-XL~\citep{nuwaxl} pursue directions that are largely \emph{orthogonal} to ours: LaVie targets high-resolution text-to-video generation via cascaded latent diffusion, VideoCrafter2~\citep{chen2024videocrafter2} addresses the data-efficiency of text-conditional training, and NUWA-XL proposes a hierarchical ``diffusion-over-diffusion'' scheme for extremely long videos. None of these methods modifies the underlying complexity of the temporal mixing layer itself. Because our temporal SSM block is a drop-in replacement for temporal self-attention, it can, in principle, be combined with the cascaded, autoregressive, or hierarchical pipelines above to yield additional gains at long sequence lengths, and a direct empirical comparison or integration with these systems is an important direction for future work.

\subsection{SSMs and Their Applications}
\label{ssm_priorworks}
Mamba~\citep{mamba}, a recent SSM, offers efficient computation and demonstrates outstanding performance.
In HiPPO~\citep{hippo} and S4~\citep{s4}, the groundwork for subsequent advancements is established in sequence modeling frameworks, including the development of Mamba. 
By introducing a structured parameterization of SSMs, S4 enabled efficient computation and demonstrated exceptional performance in capturing long-range dependencies~\citep{lra}. 
S4D~\citep{s4d} is a simplified version of S4, which introduced a diagonal matrix formulation.
SSMs have been applied across various domains, including image and video classification~\citep{visionmamba, vmamba, li2024videomamba}, language modeling~\citep{pretrainssm, h3, lieber2024jamba, zhao2024cobra}, and reinforcement learning~\citep{ssmrl1, s4wm}. 

In the field of diffusion models using SSMs, DiffuSSM~\citep{diffussm} was the first to integrate SSMs into diffusion models by replacing the computationally intensive spatial attention in image generation with SSMs.
Subsequent work has likewise focused on replacing \emph{spatial} attention with SSMs for image generation, or on handling spatio-temporal information globally without separating the two axes~\citep{hu2024zigma, fu2024lamamba, hong2025hth}.
A concurrent and independent line of work, Matten~\citep{gao2024matten}, also explores Mamba-based architectures for video generation; however, Matten combines Mamba with attention primarily along the spatial axis and targets shorter, higher-resolution clips, whereas our work explicitly targets the temporal axis with a pure bidirectional SSM and focuses on long-term ($\geq 256$-frame) generation under matched compute budgets.
This paper is an extended version of our preliminary work presented at the ICLR 2024 Workshop; the present manuscript adds the bidirectionality and selective-scan ablations, the GQN-Mazes and CARLA Town01 experiments, and the scaling analysis up to 913M parameters.

\section{Experiments}

\begin{table*}[t] \centering \caption{Model Configuration. Settings for the UNet-based diffusion models used in the experiments. The models are scaled such that the base channel size is proportional to the number of attention heads, whereas the hidden dimension is fixed at 64.} \scalebox{0.88}{ \begin{tabular}{cccccc} \toprule \multicolumn{6}{l}{\textbf{Attention-based Models}} \\ \midrule Params & \# Base Channels & \# Attention Heads & \# Attention Dims & Spatial Module & Temporal Module\\ \midrule 14.1M & 32 & 4 & 64 & Linear attn. & Attention \\ 56.3M & 64 & 8 & 64 & Linear attn. & Attention \\ 71.2M & 72 & 9 & 64 & Linear attn. & Attention \\ 225M & 128 & 16 & 64 & Linear attn. & Attention \\ 900M & 256 & 32 & 64 & Linear attn. & Attention \\ \bottomrule \toprule \multicolumn{6}{l}{\textbf{SSM-based Models}} \\ \midrule 14.5M & 32 & 4 & 64 & Linear attn. & SSM \\ 57.5M & 64 & 8 & 64 & Linear attn. & SSM \\ 175M & 112 & 14 & 64 & Linear attn. & SSM \\ 229M & 128 & 16 & 64 & Linear attn. & SSM \\ 913M & 256 & 32 & 64 & Linear attn. & SSM \\ \bottomrule \end{tabular} } \label{tab:model_comparison} \end{table*}

\begin{figure}[t]
    \centering
    \includegraphics[width=\linewidth]{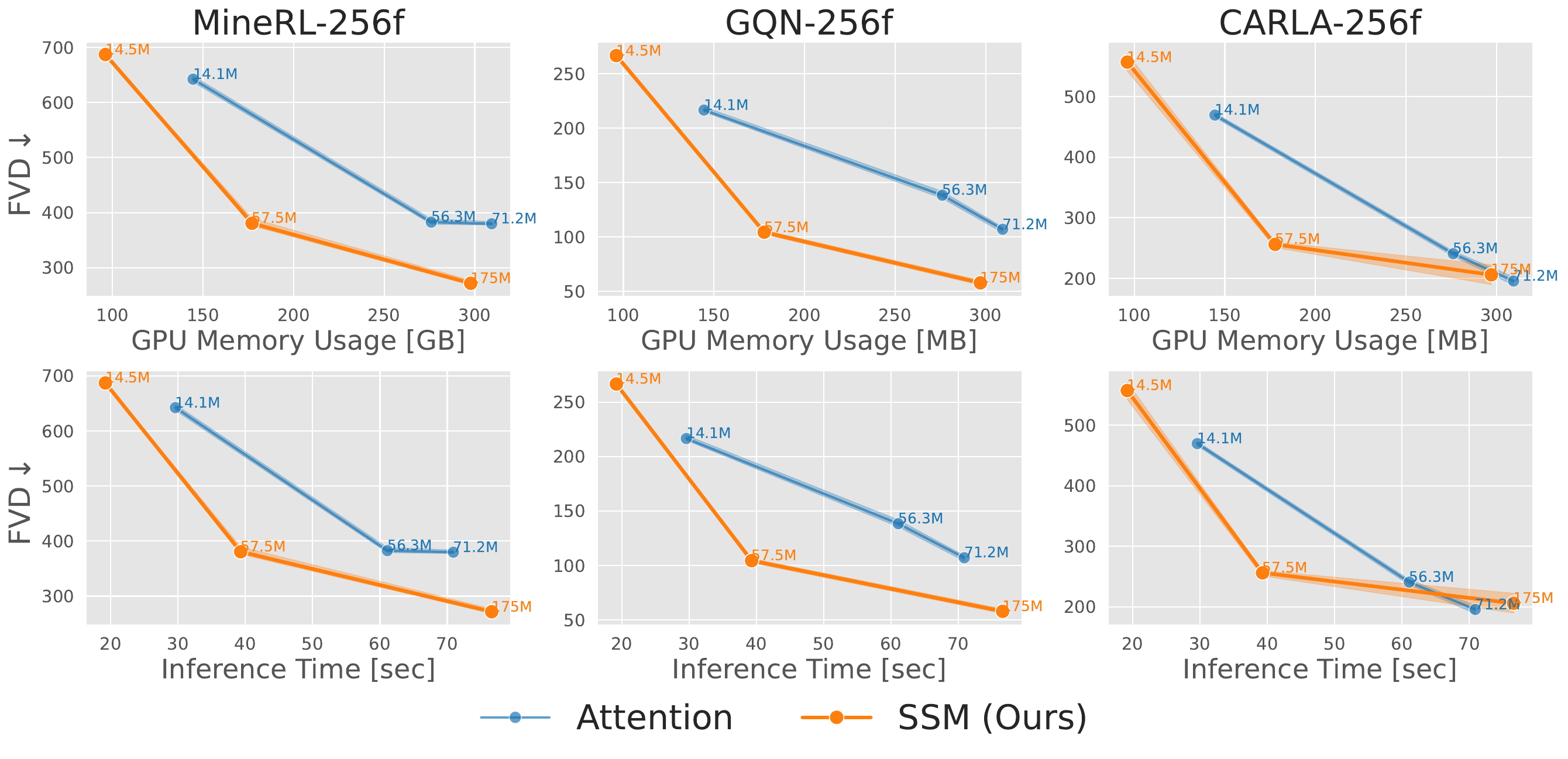}
    \caption{Comparison of GPU memory usage, inference time, and generative performance (FVD $\downarrow$) between the temporal SSM layer and temporal attention for 256 frames on the MineRL Navigate, GQN-Mazes, and CARLA Town01 datasets. Each plot point includes the model size next to it.
    The temporal SSM layer consistently demonstrates superior performance on 256-frame sequences under the same computational resource constraints.
    }
    \label{fig:combined}
\end{figure}

In this section, we present a series of experiments demonstrating that employing SSMs to capture temporal dependencies in long-duration video generation (e.g., 256 frames) yields higher quality than conventional attention-based models under the same computational constraints (e.g., memory and runtime). 
Notably, across multiple datasets and model sizes, temporal SSMs consistently outperformed in generating long-term videos under equivalent memory and computational time constraints. 
These findings suggest that incorporating SSMs into the temporal layers is advantageous for building long-term video diffusion models with limited computational resources.

\subsection{Experimental Setup}

{\bf Models} \ 
For each experimental setup, we conducted experiments using both VDMs with temporal attention layers and VDMs with SSM-based temporal layers, across multiple model sizes. 
The number of parameters and scaling methods used for each model are listed in \autoref{tab:model_comparison}.
The model size for generating 256-frame-long videos was selected based on the scaling approach used in this study, ensuring that the experiments were feasible within the constraints of the available hardware, which comprised eight NVIDIA A100 GPUs (40 GB).
The VDMs were implemented based on a publicly available and widely used implementation \footnote{\url{https://github.com/lucidrains/video-diffusion-pytorch}} that utilizes UNet's convolution layer and linear attention~\citep{shen2019efficient,kitaev2020reformer,wang2020linformer} to capture spatial features. 
In this study, we focused on comparing only the layers responsible for capturing temporal features, as changes in spatial feature extraction were not relevant.
\\
{\bf Datasets} \ 
We first used the MineRL Navigate dataset~\citep{minerl, cwvae}, a widely used dataset for long-term video generation derived from the Minecraft environment. 
The training set comprised 1,186 videos, combining both the train and test splits. 
We trained the proposed models on videos with frame lengths of 16, 64, and 256 to evaluate computational cost and generative performance when using SSMs versus attention as the temporal layer. 
In addition, we employed the GQN-Mazes~\citep{eslami2018neural, cwvae} and CARLA-Town01~\citep{dosovitskiy2017carla, fdm} datasets for long-term video generation tasks. 
The GQN-Mazes dataset contains 108,200 videos from a 3D maze simulator, whereas the CARLA Town01 dataset contains 508 videos from a driving simulator. 
For these datasets, we trained our models using videos with 256 frames each. 
Across all experiments, the spatial resolution of the videos was fixed at $32\times32$ pixels. \\
{\bf Baseline} \ 
We established the experimental baseline using VDMs~\citep{vdm}. 
Our analysis was designed to alter only the temporal attention layers in the VDMs using the proposed temporal SSM layers. 
This strategy enabled a focused examination of the impact and efficacy of the temporal SSM layers in video generation, facilitating a direct comparison with existing temporal layers. \\
{\bf Evaluation Metrics} \ 
In our validation process, we evaluated the quality of the video samples generated by the trained models. 
To evaluate the quality of the generated videos, we employed the Fréchet Video Distance (FVD) metric~\citep{fvd}, using an I3D network pretrained on the Kinetics-400 dataset~\citep{kinetics}. 
The FVD is a recognized standard for assessing the quality of generated videos~\citep{vdm, makeavideo, imagenvideo, fdm}, with lower scores indicating higher quality. 
For the MineRL Navigate dataset, the calculations involve all 1,186 videos and 1,000 generated samples. 
For the GQN-Mazes, the calculations involve 2,000 videos from the dataset and 2,000 samples, whereas for the CARLA Town01, all 508 videos from the dataset and 500 samples were used.
We used three different random seeds to introduce variations in data sampling and sample generation for the FVD calculation.

\begin{figure}[t]
    \centering
    \includegraphics[width=\linewidth]{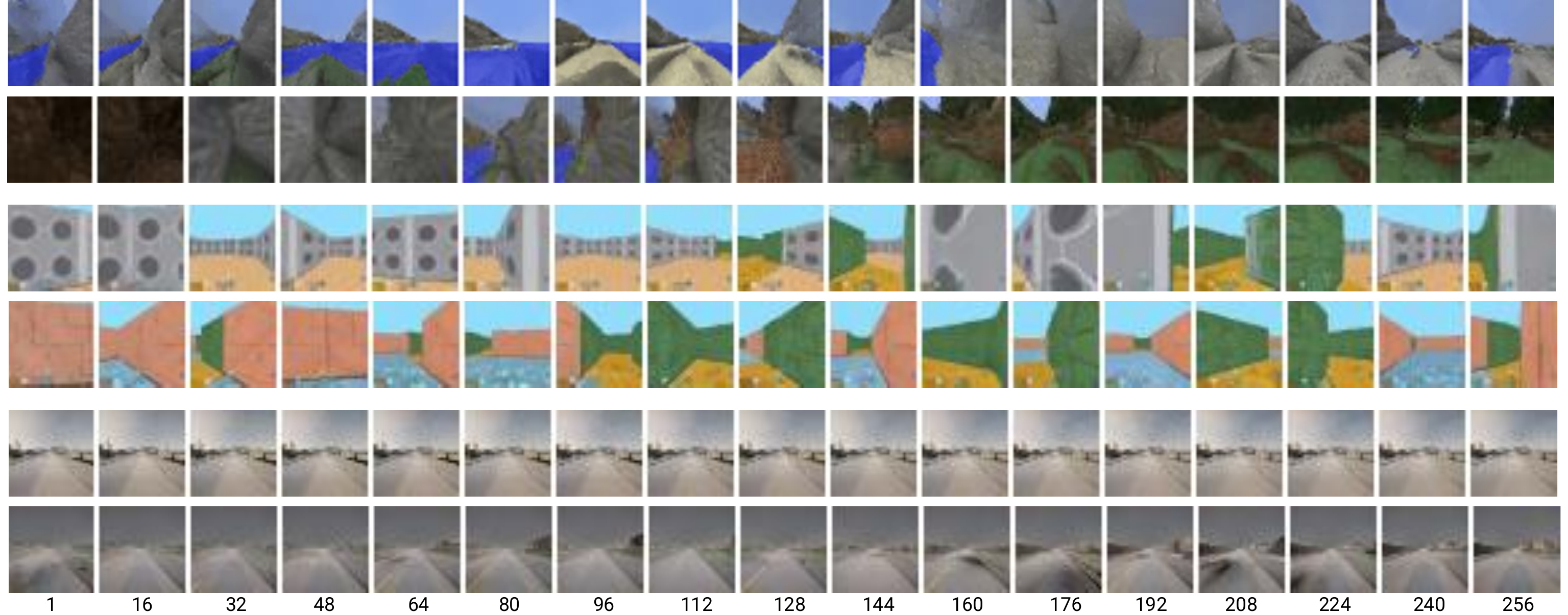}
    \caption{Videos generated by temporal SSM-based VDMs. Each row corresponds to a single generated 256-frame video (frames are displayed left to right). The top two rows are from MineRL Navigate, the middle two from GQN-Mazes, and the bottom two from CARLA Town01. 
    }
    \label{fig:minerl_qualitative}
\end{figure}

\subsection{Main Results}

We trained models of various sizes across different settings, and the results are shown in \autoref{fig:combined} and \autoref{fig:short_frames}.
First, we describe the results obtained from the MineRL Navigate dataset with various video frames. 
For sequences of 16 and 64 frames, the VDM with an SSM-based temporal layer demonstrates a generative performance comparable to that of the attention-based temporal layer. 
However, when the frame count is increased to 256 frames, the SSM-based VDM exhibits superior generative performance relative to memory usage and computational time compared with the attention-based approach.
Furthermore, experiments conducted on the GQN-Mazes and CARLA Town01 confirm the advantages of using an SSM-based temporal layer in VDMs over attention-based layers. 
The samples generated from the temporal SSM-based VDMs are shown in \autoref{fig:minerl_qualitative}.

\begin{figure}[t]
    \centering
    \includegraphics[width=0.8\linewidth]{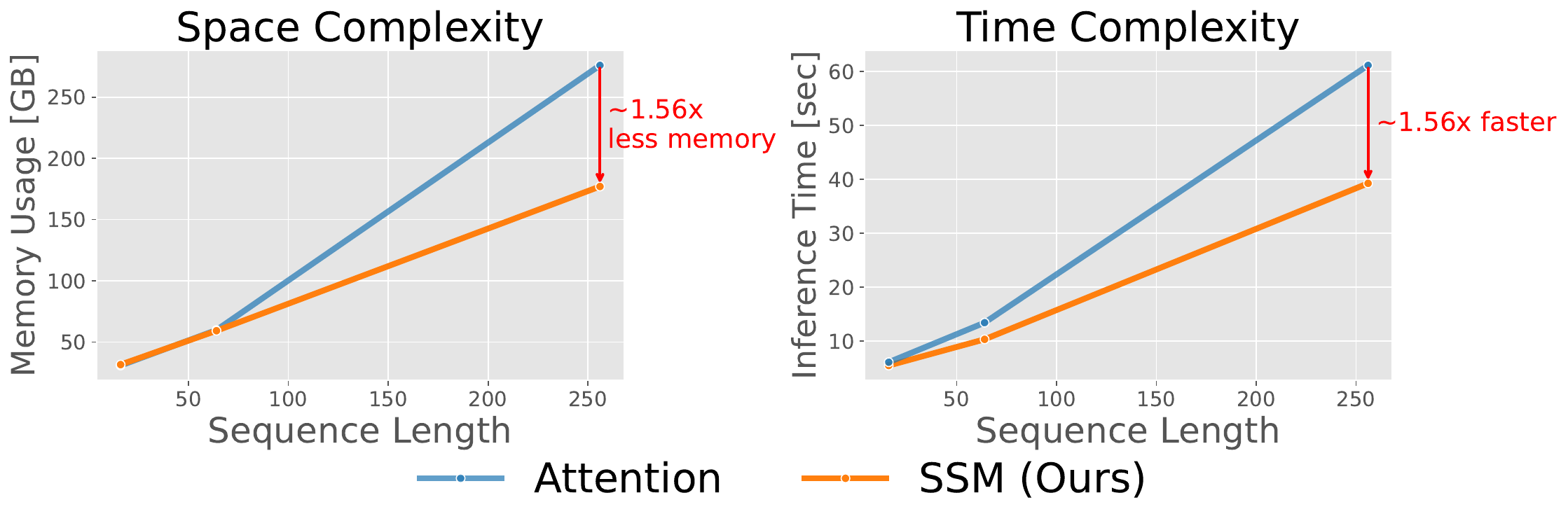}
    \caption{(\textbf{Left}): Memory consumption during training with 8 NVIDIA A100 GPUs (40 GB) at a batch size of 8. (\textbf{Right}): Inference time for generating a sample with a single NVIDIA A100 GPU. 
    }
    \label{fig:computation}
\end{figure}

\begin{figure}[t]
    \centering
    \includegraphics[width=\linewidth]{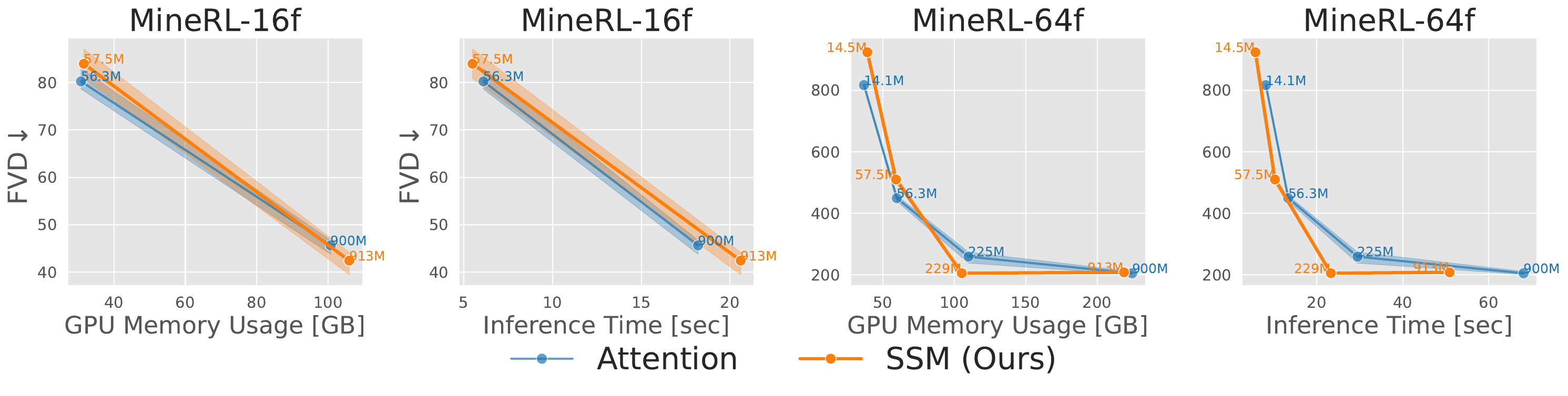}
    \caption{Comparison of GPU memory usage, inference time, and generative performance (FVD$\downarrow$) on 16 frames (\textbf{Left}) and 64 frames (\textbf{Right}) for the MineRL Navigate dataset.
    In contrast to the 256-frame experiment (\autoref{fig:minerl_qualitative}), both the temporal SSM layer and the temporal attention mechanism exhibit comparable performance on shorter sequences.
    }
    \label{fig:short_frames}
\end{figure}

In \autoref{fig:computation}, we show the variation of the training memory usage and inference time with the video sequence length for each temporal layer using the same model size. 
Memory consumption is measured on a batch size of eight, while inference times reflect sample generation on a single NVIDIA A100 GPU.
The SSM-based model achieved approximately a $\times 1.56$ improvement in inference speed and a similar reduction in memory usage at 256 frames, highlighting its efficiency compared to attention-based approaches under the same model size.
In addition, the results demonstrate that, whereas memory usage and inference speed in attention-based models increase quadratically with sequence length, the SSM-based model exhibits only a linear increase.

While the results indicate that SSMs can serve as viable alternatives to attention mechanisms in the temporal layers for relatively short videos (e.g., 16 or 64 frames), their superiority becomes more pronounced when generating longer videos (e.g., 256 frames), where SSM-based models consistently yield higher-quality outputs compared to attention-based models. 
This improvement stems from SSMs’ ability to accommodate larger model sizes while retaining comparable computational resource requirements.
Additionally, based on the observed relationship between the number of frames and computational cost, this trend is expected to become more pronounced as the number of video frames increases.

\subsection{Ablation Study of Temporal SSM Layers}
\label{ablation_study}


In our ablation study, we investigated the impact of bidirectional processing and selective scans on the generation of high-quality videos, particularly in capturing temporal features.
Focusing on 256-frame video generation from the MineRL Navigate dataset, we observed that employing a unidirectional SSM or an unselective scanning approach~\citep{s4d} markedly degraded generative quality (\autoref{tab:unidirectional_ablation}).
In the spatial domain, SSM-based diffusion models have demonstrated performance improvements by processing information in multiple directions~\citep{diffussm, hu2024zigma, fu2024lamamba}, an effect attributable to the fundamentally causal nature of SSMs.
Consistent with these previous findings, our ablation results indicate that relaxing purely causal constraints yields similar benefits in the temporal domain.

Beyond bidirectionality, our results also underscore the importance of selective scans over the unselective S4D parameterization~\citep{s4d}. We interpret this gap through the input-dependent gating intrinsic to selective SSMs~\citep{mamba}: since $\mathbf{B}$, $\mathbf{C}$, and the discretization step $\mathbf{\Delta}$ are functions of the input at each frame (Algorithm~\ref{alg:temporal_ssm}, lines~5--7), a small $\mathbf{\Delta}_t$ effectively suppresses the state update at frame $t$ --- preserving the hidden state across redundant or low-information frames --- whereas a large $\mathbf{\Delta}_t$ allows salient frames to drive state transitions. In contrast, unselective SSMs such as S4D apply a fixed, input-independent transition and cannot perform such frame-wise gating. Empirically, this distinction becomes more pronounced as the sequence grows: redundancy is greater in 256-frame videos than in short clips, which is consistent with the substantial FVD gap we observe between selective and unselective variants in the long-sequence regime. While direct visualization of $\mathbf{\Delta}$ across frames would provide further insight into this mechanism, a comprehensive analysis of learned $\mathbf{\Delta}$ patterns and their correspondence to visually salient transitions is left to future work.

\begin{table*}[t]
    \centering
    \caption{Ablation study of temporal SSM layers. Employing bidirectionality and selective scans enhances video generation performance.}
    \scalebox{0.90}{
    \begin{tabular}{cccccc}
        \toprule
        \textbf{Bidirectional} & \textbf{Selective Scan} & Params & Memory Usage$\downarrow$ & Inference Time$\downarrow$ & FVD $\downarrow$ \\
        \midrule
        \textcolor{cb_red}{\XSolidBrush} & \textcolor{cbgreen}{\textbf{\Checkmark}} & 56.9 M & 146 GB & 31.8 sec. & 610 \\
        \textcolor{cbgreen}{\textbf{\Checkmark}} & \textcolor{cb_red}{\XSolidBrush} & 56.3 M & 200 GB & 37.3 sec. & 458 \\
        \textcolor{cbgreen}{\textbf{\Checkmark}} & \textcolor{cbgreen}{\textbf{\Checkmark}} & 57.5 M & 177 GB & 39.3 sec. & \bf{381} \\
        \bottomrule
    \end{tabular}
    }
    \label{tab:unidirectional_ablation}
\end{table*}



\section{Discussion}

\subsection{Future directions.}
This study opens multiple research directions.
Integrating temporal SSM layers with Latent Diffusion Models~\citep{ldm, lvdm, magicvideo} can enhance video generation at higher spatial resolutions (e.g., up to $8\times$) without substantially increasing computational cost. 
Moreover, this approach can be combined with distillation~\citep{song2023consistency, heek2024multistep, li2024t2vturbo} to accelerate inference and improve computational efficiency. 
Although we focused on unconditional video generation, the proposed method can be extended to conditional tasks, such as guidance~\citep{classifierguidance, cfg} and reward alignment~\citep{oshima2025inference}.
Beyond video generation, similar quadratic-complexity bottlenecks of attention also arise in other settings, such as multi-reference image generation~\citep{openai2025gpt4oimage, google2025nanobanana, wu2025qwenimagetechnicalreport}, where the cost grows with the number of reference images~\citep{oshima2025multibanana}; extending our SSM-based design to such settings is a promising direction for future work.

\subsection{Limitations and scope.}
We deliberately fixed the spatial resolution at $32\times32$ across all experiments in order to \emph{isolate} the effect of the temporal layer; under this controlled setting, any difference in generation quality between attention- and SSM-based variants can be attributed to the temporal mixing mechanism itself rather than to confounding spatial design choices. Scaling to higher resolutions (e.g., $\geq 128\times128$) is a natural next step and we expect it to be most efficiently realized by combining temporal SSM layers with a latent diffusion backbone~\citep{ldm, lvdm, magicvideo} or with cascaded super-resolution stages~\citep{imagenvideo}, both of which are orthogonal to the contribution of this paper.

Several practical constraints also remain when scaling beyond 256 frames. First, although SSMs achieve linear-time complexity in $L$, activation memory still grows linearly with sequence length, so generating substantially longer videos at our model sizes would require activation checkpointing or model parallelism. Second, large-scale long-form video datasets suitable for unconditional training remain scarce. Third, FVD evaluation becomes less reliable at very long horizons, since the I3D backbone used by FVD was pretrained on short clips and may not faithfully reflect long-range temporal coherence; developing evaluation protocols suitable for $\geq 512$-frame sequences is itself an open problem. Addressing these data and evaluation issues is a prerequisite for meaningful empirical analysis beyond 256 frames.

Finally, our evaluation relies solely on FVD, a standard but imperfect automatic metric for video generation. Complementing FVD with human perceptual studies, particularly to assess temporal coherence and long-range consistency that automatic metrics may overlook, is an important direction for future work.

\section{Conclusion}
This paper demonstrates that incorporating SSMs into the temporal layers of diffusion models for video generation yields superior long-term video modeling while maintaining the same computational budget (e.g., memory usage, computational time). 
These findings underscore the adaptability of temporal SSM layers in advancing video diffusion models and highlight their extensive potential for future developments in the field. 
Moreover, this approach enhances access to cutting-edge video generation research, enabling institutions with limited computational resources to participate and potentially accelerate progress and innovation.

\clearpage
\section*{Acknoledgement}
\textbf{Availability Statement}~~The experiment code for this study is publicly available. 
The datasets used for the experiments in this study are publicly available through the Internet.

\noindent
\textbf{Funding}~~This work was supported by the Japan Society for the Promotion of Science (JSPS) KAKENHI Grant Number J23H04974. 


\bibliography{paper}
\bibliographystyle{iclr2026_conference}

\clearpage
\appendix

\section{Experimental Settings}
\label{training_details}
To ensure a fair comparison of the modules extracting the temporal relationships under the same resolution settings, all configurations except for the temporal layers, were the same in our experiments. 
We used NVIDIA A100 $\times 8$ (from a cloud provider). 
Hyperparameters, which are consistent across all model sizes and datasets, are shown in \autoref{tab:hyparam}.

\begin{table*}[h]
    \centering
    \caption{Hyperparameters used in the experiments.}
    \begin{tabular}{l|l}
        \hline
        \textbf{Parameter} & \textbf{Value} \\
        \hline
        UNet channel multipliers & (1, 2, 4, 8) \\
        Time embedding dimension & 1024 \\
        Time embedding linears & 2 \\
        Denoising timesteps ($T$) & 256 \\
        Loss type & L2 loss of noise $\epsilon$ \\
        Training steps & 100k \\
        Optimizer & Adam ($\beta_1=0.9$, $\beta_2=0.999)$ \\
        Learning rate & 0.00001 \\
        Train batch size & 8 \\
        EMA decay & 0.9999 \\
        \hline
    \end{tabular}
    \label{tab:hyparam}
\end{table*}

\section{Additional Qualitative Results}
\begin{figure}[ht]
    \centering
    \includegraphics[width=\linewidth]{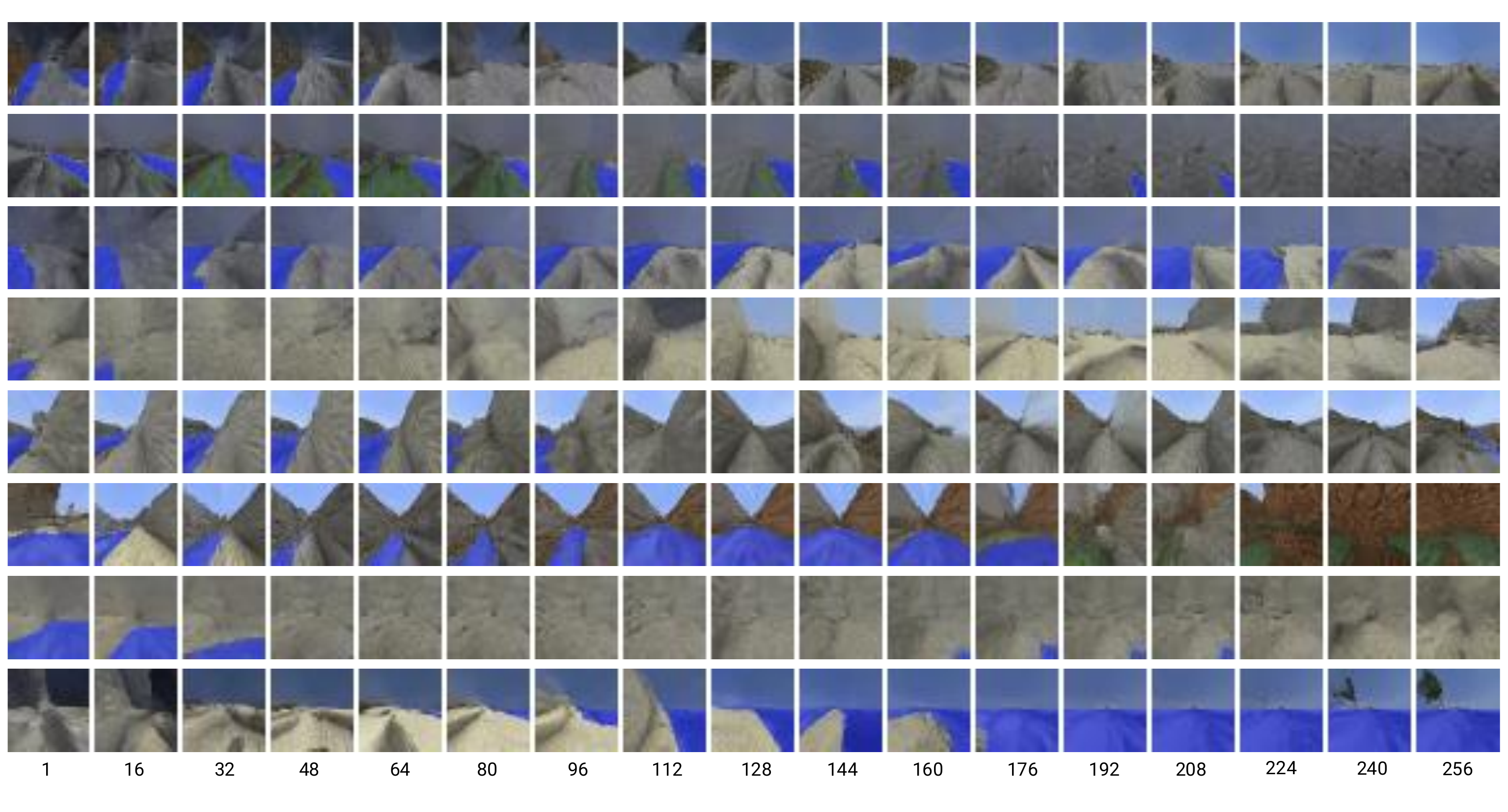}
    \caption{Additional qualitative generation results in MineRL Navigate (\# of frames are 256). Each column represents different samples.}
    \label{minerl_qualitative}
\end{figure}

\begin{figure}[ht]
    \centering
    \includegraphics[width=\linewidth]{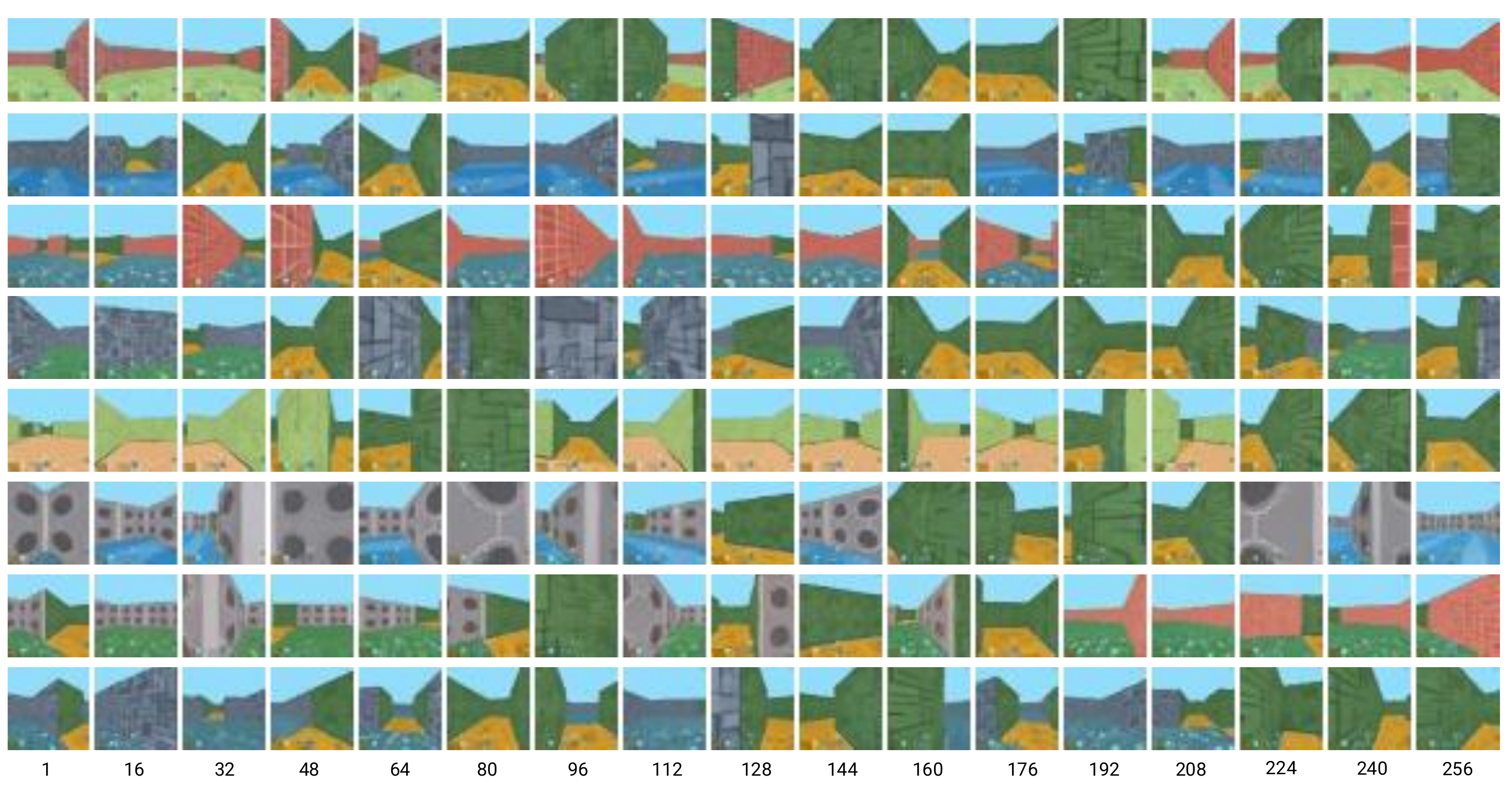}
    \caption{Additional qualitative generation results in GQN-Mazes (\# of frames are 256). Each column represents different samples.}
    \label{gqn_qualitative}
\end{figure}

\begin{figure}[ht]
    \centering
    \includegraphics[width=\linewidth]{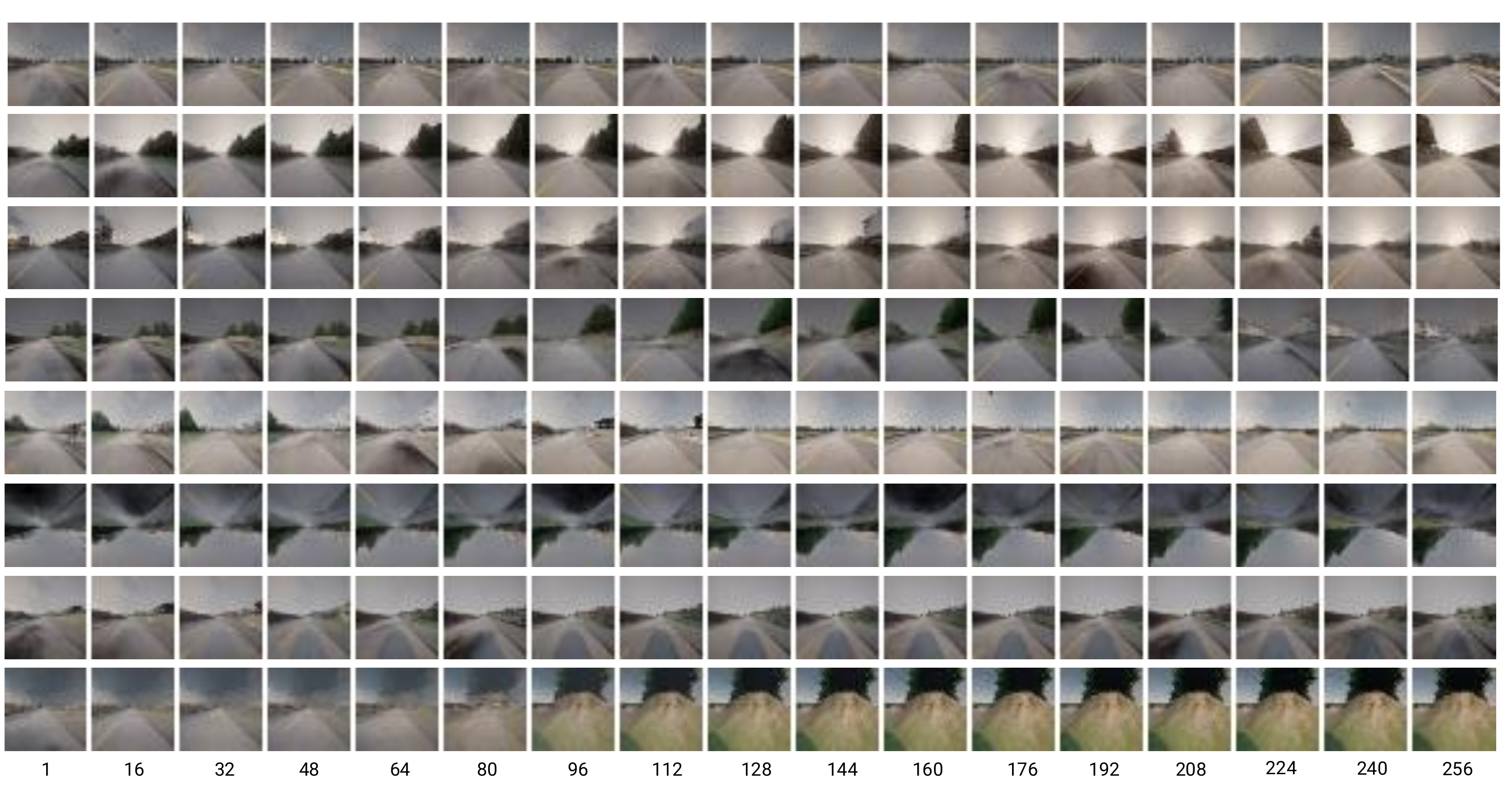}
    \caption{Additional qualitative generation results in CARLA Town01 (\# of frames are 256). Each column represents different samples.}
    \label{carla_qualitative}
\end{figure}

Shown in \autoref{minerl_qualitative},~\autoref{gqn_qualitative},~\autoref{carla_qualitative}.
\label{additional_qualitative}

\clearpage
\section{Quantitative Results}
Shown in \autoref{tab:quantitative}, \autoref{tab:quantitative_short} in table form.

\begin{table*}[ht]
    \centering
    \caption{Quantitative results of experiments. Memory consumption during training with 8 NVIDIA A100 GPUs (40 GB) at a batch size of 8. Inference time for generating a sample with a single NVIDIA A100 GPU.}
    \begin{tabular}{lccc}
        \toprule
        \multicolumn{3}{l}{\textbf{MineRL Navigate 256 frames}} \\
        Models & Memory Usage$\downarrow$ & Inference Time$\downarrow$ & FVD$\downarrow$ \\
        \hline
        Attention-14.1M & 144 GB  & 29.6 sec.  & 642 \\
        Attention-56.3M & 276 GB  & 61.1 sec.  & 383 \\
        Attention-71.2M & 309 GB  & 70.8 sec.  & 380 \\
        \midrule
        SSM-14.5M (Ours)     & 96.1 GB  & 19.2 sec.  & 687 \\
        SSM-57.5M (Ours)     & 177 GB  & 39.3 sec.  & 381 \\
        SSM-175M (Ours)      & 298 GB & 76.6 sec. & 272 \\
        \bottomrule
        \toprule
        \multicolumn{3}{l}{\textbf{GQN-Mazes 256 frames}} \\
        Models & Memory Usage$\downarrow$ & Inference Time$\downarrow$ & FVD $\downarrow$ \\
        \hline
        Attention-14.1M & 144 GB  & 29.6 sec.  & 217 \\
        Attention-56.3M & 276 GB  & 61.1 sec.  & 138 \\
        Attention-71.2M & 309 GB  & 70.8 sec.  & 107 \\
        \midrule
        SSM-14.5M (Ours)     & 96.1 GB  & 19.2 sec.  & 267 \\
        SSM-57.5M (Ours)     & 178 GB  & 39.3 sec.  & 104 \\
        SSM-175M (Ours)      & 297 GB & 76.6 sec. & 57.8 \\
        \bottomrule
        \toprule
        \multicolumn{3}{l}{\textbf{CARLA Town01 256 frames}} \\
        Models & Memory Usage$\downarrow$ & Inference Time$\downarrow$ & FVD $\downarrow$ \\
        \hline
        Attention-14.1M & 145 GB  & 29.6 sec.  & 470 \\
        Attention-56.3M & 276 GB  & 61.1 sec.  & 241 \\
        Attention-71.2M & 309 GB  & 70.8 sec.  & 196 \\
        \midrule
        SSM-14.5M (Ours)     & 96.1 GB  & 19.2 sec.  & 557 \\
        SSM-57.5M (Ours)     & 178 GB  & 39.3 sec.  & 256 \\
        SSM-175M (Ours)      & 297 GB & 76.6 sec. & 206 \\
        \bottomrule
    \end{tabular}
    \label{tab:quantitative}
\end{table*}

\begin{table*}[ht]
    \centering
    \caption{Quantitative results of experiments. Memory consumption during training with 8 NVIDIA A100 GPUs (40 GB) at a batch size of 8. Inference time for generating a sample with a single NVIDIA A100 GPU.}
    \begin{tabular}{lccc}
        \toprule
        \multicolumn{3}{l}{\textbf{MineRL Navigate 16 frames}} \\
        Models & Memory Usage$\downarrow$ & Inference Time$\downarrow$ & FVD$\downarrow$ \\
        \midrule
        Attention-56.3M & 30.8 GB  &  6.11 sec. & 80.3 \\
        Attention-900M  & 101 GB & 18.2 sec. & 45.6 \\
        \midrule
        SSM-57.5M (Ours)  & 31.6 GB  & 5.49 sec. & 84.0 \\
        SSM-913M (Ours)   & 106 GB & 20.6 sec. & 42.4 \\
        \bottomrule
        \toprule
        \multicolumn{3}{l}{\textbf{MineRL Navigate 64 frames}} \\
        Models & Memory Usage$\downarrow$ & Inference Time$\downarrow$ & FVD$\downarrow$ \\
        \midrule
        Attention-14.1M & 36.8 GB  & 8.19 sec.  & 817 \\
        Attention-56.3M & 59.8 GB  & 13.4 sec.  & 449 \\
        Attention-225M  & 110 GB & 29.5 sec. & 259 \\
        Attention-900M  & 224 GB & 68.1 sec. & 205 \\
        \midrule
        SSM-14.5M (Ours)     & 36.8 GB  & 5.76 sec.  & 923 \\
        SSM-57.5M (Ours)     & 59.2 GB  & 10.3 sec.  & 510 \\
        SSM-229M (Ours)      & 105 GB & 23.3 sec. & 205 \\
        SSM-913M (Ours)      & 219 GB & 50.9 sec. & 208 \\
        \bottomrule
    \end{tabular}
    \label{tab:quantitative_short}
\end{table*}

\end{document}